\pdfoutput=1

\documentclass[11pt]{article}

\usepackage{ACL2023}

\usepackage{times}
\usepackage{latexsym}

\usepackage[T1]{fontenc}

\usepackage[utf8]{inputenc}

\usepackage{microtype}

\usepackage{inconsolata}
\usepackage{graphicx}

\title{Beyond Benchmarking: A New Paradigm for Evaluation and Assessment of Large Language Models}

\author{Jin Liu \and Qingquan Li \and Wenlong Du \\
        Independent Researchers}

\begin{document}
\maketitle
\begin{abstract}
In current benchmarks for evaluating large language models (LLMs), there are issues such as evaluation content restriction, untimely updates, and lack of optimization guidance.
In this paper, we propose a new paradigm for the measurement of LLMs: Benchmarking-Evaluation-Assessment. Our paradigm shifts the ``location'' of LLM evaluation from the ``examination room'' to the ``hospital''.
Through conducting a ``physical examination'' on LLMs, it utilizes specific task-solving as the evaluation content, performs deep attribution of existing problems within LLMs, and provides recommendation for optimization.
\end{abstract}

\section{Introduction}

Currently, research on large language models (LLMs) has become a hot topic.
In order to measure the performance of LLMs, a large number of studies on evaluation benchmarks of LLMs have been spawned.
Those includes general evaluation benchmarks such as SuperCLUE~\cite{xu2023superclue} and MMBench~\cite{liu2023mmbench}, domain knowledge evaluation datasets such as Xiezhi~\cite{gu2024xiezhi} and FinEval~\cite{zhang2023fineval}, and safety and hallucination evaluation methods such as SafetyPrompts~\cite{sun2023safety} and HaluEval~\cite{li2023halueval}.
Based on this, each LLM has a wealth of benchmarks to measure its own performance.
However, in the promotion of LLMs, we often see such titles: ``Surpassing GPT-4, we ranks first in the capability of ...''
Many developers say that their LLMs are the best in some certain abilities, which makes us re-examine the current LLM benchmarking paradigm:
Can a batch of static data and a score be used to accurately and completely measure the performance of a LLM?

\section{Potential Issues in Current Paradigm of Benchmarking LLMs}

\subsection{Limited Benchmarking Capability: Knowledge}

Generative language models can finish given tasks through natural language dialogue.
Therefore, most benchmarks of LLMs adopt the common evaluation form in current human society - exams, that is, the knowledge to be evaluated is written into questions for LLMs to answer.
For instance, a SuperCLUE's example of evaluation on semantics understanding ability of LLMs is as follows:
\begin{itemize}
    \item \textit{Question: In the middle of the party, a Chinese man said to a foreigner: ``I'm going to `Fangbian'.'' What does ``Fangbian'' mean here?}
    \item \textit{Assistant A's Answer: In this context, ``Fangbian'' means to go to the bathroom or toilet. The Chinese man say this to politely tell the foreigner that he need to leave for a while to go to the bathroom. This is a euphemism to avoid directly mentioning the topic of going to the bathroom.}
    \item \textit{Assistant B's Answer: In this context, ``Fangbian'' means that the Chinese man need to go to the toilet or take care of other personal needs and need to leave for a while. This is a common polite expression to indicate that he is leaving temporarily without affecting the party.}
\end{itemize}

Especially in the benchmarks of evaluating professional LLMs, many benchmarks directly use professional qualification examinations, such qualification exam for legal professionals, certification of securities professional, and medical licensing examination to evaluate the ability of a certain professional field of LLMs.
An objective score can be given through exams.
However, just like students who graduate will eventually enter the workplace, LLMs also will be applied in a certain scenario.
Can the exam truly evaluate the ability of LLMs to solve professional problems?
For example, how to evaluate whether the contract written by LLMs is usable?
How to evaluate whether the investment advice by LLMs is reasonable?
The knowledge evaluation benchmark based on exams for LLMs needs to be upgraded to evaluate the ability of LLMs to solve certain professional problems.

\subsection{Evaluation Datasets Lack Dynamic Updating}

Information in the real world is constantly changing.
Based on this, developers will also update the training set to put new knowledge into LLMs.
Some existing benchmarks are periodically upgraded, but most evaluation datasets have not been updated after being released.
However, in many scenarios, new information is constantly pouring in, which required that the upgrade speed of evaluation on LLMs must keep up with the update of information.
For example, in security scenarios, new sensitive events may occur everyday.
It is necessary to update the security evaluation data timely to measure whether the LLM will generate unsafe responses.
If the safety red line is crossed, the application of LLMs will be blown.
Current dynamic update methods include:
crawling of public data, which cannot guarantee the comprehensivceness of updates;
feedback from online problems, which is a post-hoc method.
Therefore, it is necessary to discover the problems of LLMs in advance and comprehensively through dynamically updated evaluations, so as to reduce online problems and improve user experience.

\subsection{Inadequacy of Evaluation Metrics for Guiding Model Optimization}

For wrong questions in the exam, students often record the wrong questions, learn the corresponding knowledge, and do similar questions to fill in the gaps.
The purpose of doing this is to avoid the same type of mistakes in the next exam.
However, for LLMs, the goal of evaluation should not be to improve the performance on a certain evaluation metric, but to dis cover the problems of LLMs in different dimensions, and the improve the ability to solve real problems.
Moreover, there is also the risk of data leakage when training LLMs on exam questions.
Many existing evaluation benchmarks only generate a capability score, which makes it difficult ot provide targeted guidance on which abilities the LLM should improve.

\section{Benckmarking-Evaluation-Assessment: A New Paradigm for Measuring the Capability Level of LLMs}

In view of the above problems, we can evaluate the LLMs dynamically and beyond knowledge, and attribute the problems of LLMs in a targeted way, which can be compared to the process of a man discovering health problems and receiving diagnoses through physical examinations in hospitals:
\begin{enumerate}
    \item First, he undergo a comprehensive physical examination. The content of the physical examination covers most organs of his body, but the granularity of the examination is relatively coarse. Here we assume that his blood pressure is high.
    \item Secondly, in response to the blood pressure problem of the examinee, the hospital prescribes more targeted examinations, including 24-hour dynamic blood pressure, four items of standing blood pressure, electrocardiogram, cardiac color ultrasound, etc., to explore the reasons for the high blood pressure of the examinee through fine-grained metrics.
    \item Combined with fine-grained evaluation metrics, doctors diagnose the cause of the patient's illness (mental stress, primary hypertension, secondary hypertension, etc.) and provide treatment plans.
\end{enumerate}

\begin{figure*}[t]
  \centering
  \includegraphics[width=0.8\textwidth]{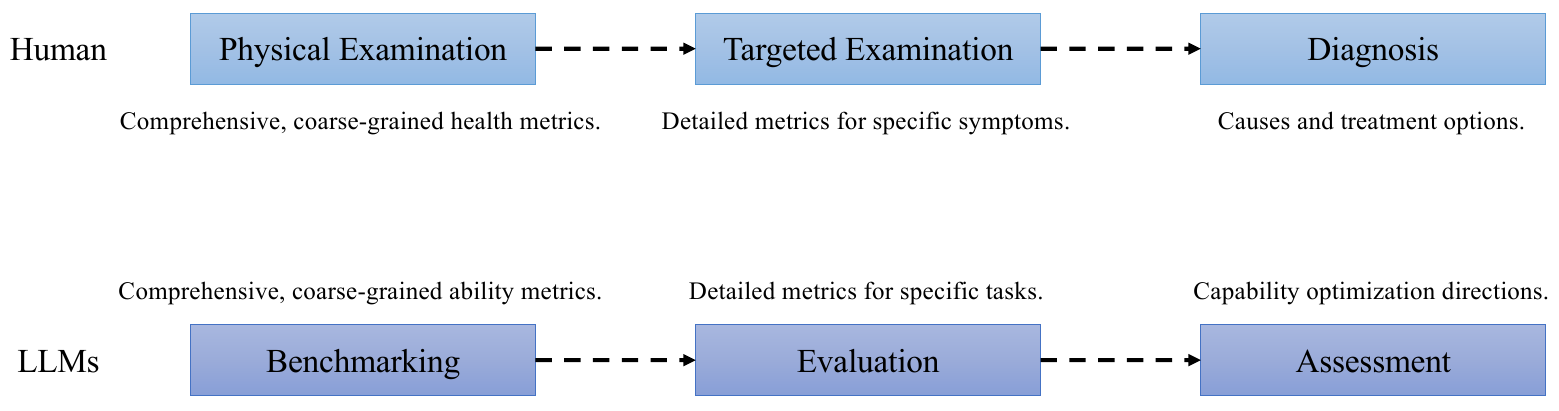}
  \caption{The comparison of measurement on human health and LLM's ability.}
  \label{fig:comp}
\end{figure*}

Referring to the medical treatment process, the ability level of LLMs can also be measured in three progressive stages: benchmarking-evaluation-assessment.
\begin{enumerate}
    \item Benchmarking: A comprehensive and coarse-grained score is conducted for the LLM to find out the lack of ability of the LLM.
    \item Evaluation: Based on the lacked capabilities, the LLM is evaluated by finishing professional tasks to further explore the specific problems of the LLM in this capability.
    \item Assessment: The problem attribution of the LLM in this capability is given through a ``doctor model'' combined with the fine-grained metrics, to providing direction for th ability improvement of the LLM.
\end{enumerate}
Fig.~\ref{fig:comp} shows the comparison of measurement on human health and LLM's ability.

\begin{figure*}[t]
  \centering
  \includegraphics[width=0.8\textwidth]{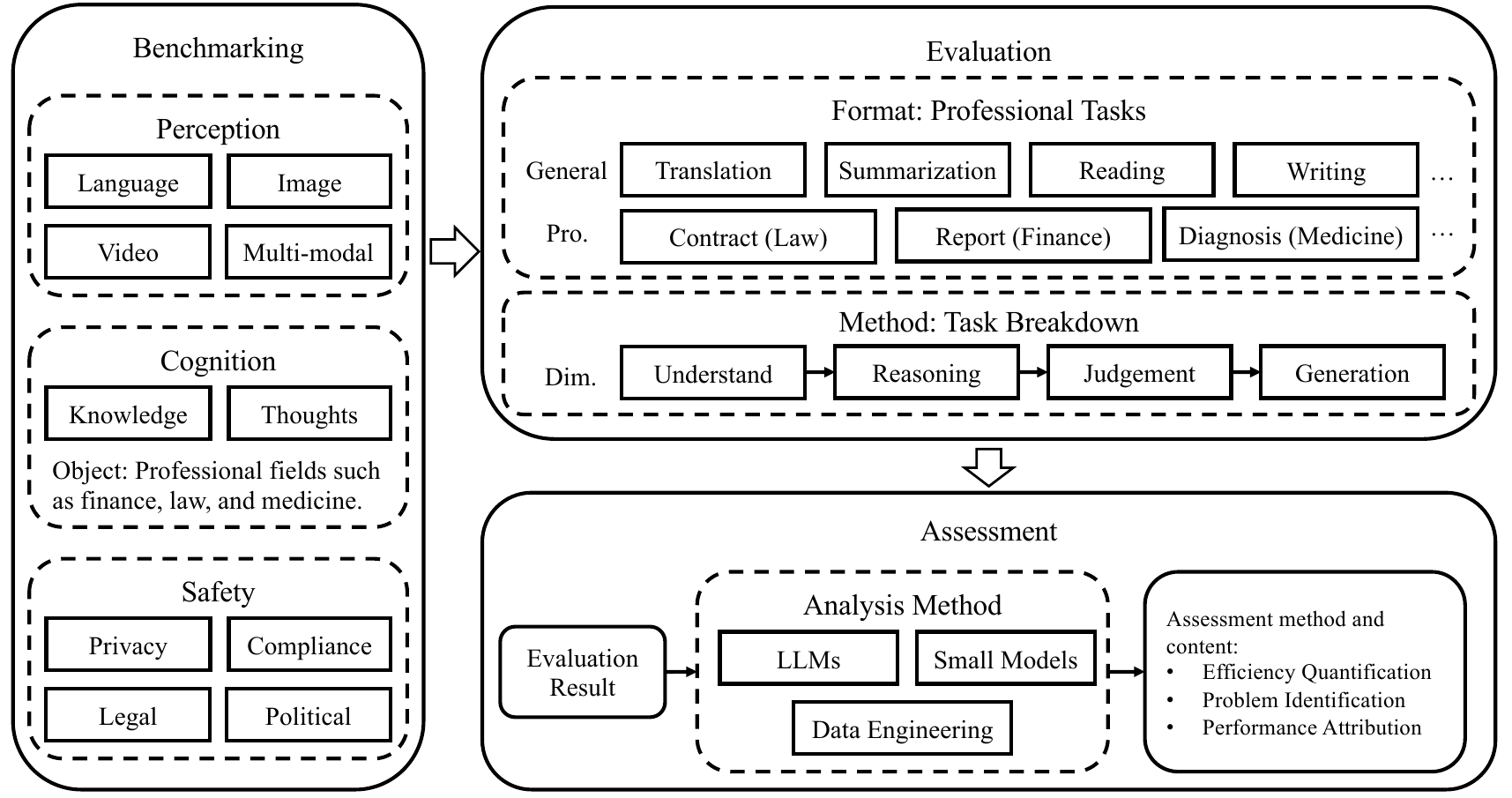}
  \caption{The architecture of our benchmarking-evaluation-assessment paradigm.}
  \label{fig:frame}
\end{figure*}

There is a progressive relationship between benchmarking, evaluation, and assessment.
Benchmarking is only a basic check of the LLM.
It uses an objective score to find problems in a certain capability of the LLM, but does not draw conclusions.
Evaluation is the core of our paradigm.
Just as the doctor can prescribe the most targeted examinations for patients based on their knowledge and experience, evaluation is a way to deeply measure the level of a LLM on a certain capability.
Through step-by-step and hierarchical measurement of different tasks, evaluation explores the capabilities of LLMs in a fine-grained manner.
Assessment, as the last stage of out paradigm, is to conduct a analysis of the results of the benchmarking and evaluation to draw conclusions.
Specifically, it is mainly to attribute the problems existing in the capabilities of LLMs and provide guidance for optimization and solutions, similar to the process of doctors prescribing medicine.
Assessment is mainly a data engineering task, which can also be executed by AI models.
Current works such as PandaLM~\cite{wangpandalm} and CritiqueLLM~\cite{ke2023critiquellm} are exploring the capabilities of LLMs in assessment.
Fig.~\ref{fig:frame} shows the preliminary architecture of our paradigm.
Each part of it needs to be designed in depth according to different scenarios and tasks.
Currently, we provide a high-level framework.

Our proposed LLM measurement paradigm benchmarking-evaluation-assessment can alleviate the potential problems or current LLM evaluation benckmarks, as follows:
\begin{enumerate}
    \item The process from benchmarking to evaluation is an expansion from traditional knowledge exam to the evaluation of task-solving ability, breaking the limitation of current LLM benchmarks being confined to knowledge.
    \item The iteration cycle of knowledge is often long, while the evaluation based on task-solving is more conducive to the dynamic update of measuring the capability of LLMs through the transformation of scenarios and tasks.
    \item In the assessment of LLMs, a doctor-like role is introduced to attribute the problems of the LLM on a specific capability, thereby providing optimization direction for the LLM on that capability.
\end{enumerate}

Visually, the above paradigm upgrade can be seen as the transformation from the LLM entering the ``examination room'' to entering the ``hospital''.
The former uses a test paper to measure the LLM's mastery of knowledge, while the feedback is only to the supplement of knowledge.
The latter locates the shortcomings of the LLM through a comprehensive physical examination, explores the roots causes of the LLM's lack of capabilities, and provides a diagnosis and treatment plan for the LLM through doctor models, which can truly discover and ``cure'' the LLM's problems.

\section{Conclusion}

As mentioned above, the current ``exam paper'' for LLMs is knowledge-based evaluation.
When the measurement paradigm of LLMs changes from ``examination room'' to ``hospital'', how to issue a ``checklist'' for LLMs, that is, what dimensions of LLMs should be evaluated, is an important topic.
The LLM is not just a knowledge center like a search engine, but a tool to improve productivity.
Therefore, the ability of a LLM to solve specific tasks should be the goal of LLM measurement.
It is necessary to decompose the task-solving process of the LLM as the evaluation dimension.
For example, to what extent does the LLM master knowledge?
How does the LLM internalize extrenal knowledge?
How does the LLM solve tasks such as language understanding, logical reasoning, and text generation step by step?
Those capability dimensions will serve as a design guide for evaluation dimensions and metrics.
We will demonstrate how to split the capability dimension of the LLM to design the evaluation framework.
Furthermore, for the assessment, how to mine and attribute problems of the LLM automatically through different methods such as data engineering and models is the last step of giving a ``checklist''.

\bibliography{anthology,custom}
\bibliographystyle{acl_natbib}




\end{document}